\documentclass{article}

\PassOptionsToPackage{sort&compress}{natbib}
\usepackage[preprint]{corl_2026} 

\usepackage{graphicx}
\usepackage{booktabs}
\usepackage{amsmath}
\usepackage{amssymb}
\usepackage{algorithm}
\usepackage{algpseudocode}
\usepackage{xcolor}
\usepackage{float}
\usepackage[most]{tcolorbox}
\usepackage{enumitem}
\usepackage{subcaption}
\newtcolorbox{promptbox}[1][]{
  breakable,
  colback=gray!10,
  colframe=gray!80,
  fontupper=\ttfamily\small,
  title=#1,
  fonttitle=\bfseries
}

\newcommand{\tab}{\hspace*{2em}}

\usepackage{xspace}

\newcommand{\ours}{\textsc{Flex}\xspace}

\newcommand{\correction}{\ensuremath{\ell}\xspace}
\newcommand{\correctionk}{\ensuremath{\ell^k}\xspace}
\newcommand{\featureset}{\ensuremath{\Phi}\xspace}
\newcommand{\featuresetcurr}{\ensuremath{\Phi^k}\xspace}

\title{Flexible Language-Guided Representation Editing for Adaptive Robot Behavior}

\author{
  Audrey Lee$^1$\\
  \texttt{\small audreylee@alum.mit.edu} \\
  \And
  Nathaniel Dennler$^1$ \\
  \texttt{\small dennler@mit.edu} \\
  \And
  Ilia Sucholutsky$^2$ \\
  \texttt{\small isucholu@purdue.edu} \\
  \And
  Andreea Bobu$^1$ \\
  \texttt{\small abobu@mit.edu} \\
}

\begin{document}
\maketitle
\vspace{-3em}
\begin{center}
  $^1$ MIT Computer Science \& Artificial Intelligence Laboratory \quad $^2$ New York University \\
\end{center}
\vspace{2em}

\begin{figure}[h!]
    \centering
    \vspace{-2em}
    \includegraphics[width=\textwidth]{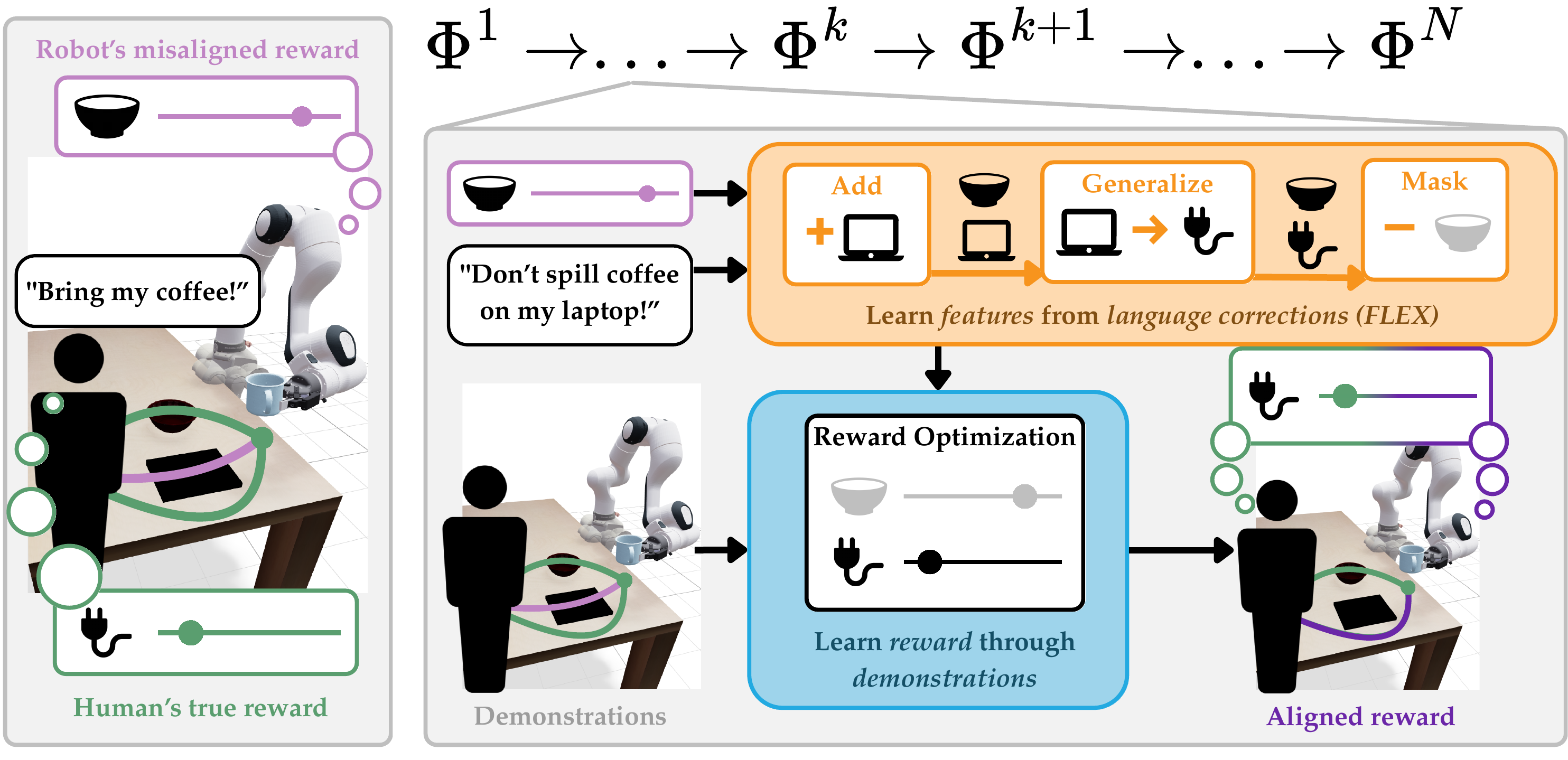}
    \caption{\ours overview. A human's task instruction (left) results in a misalignment between the human's (green) and robot's (lavender) internal task representations. To address this misalignment, \ours receives natural language corrections and applies three operations (orange) to \textsc{Add} missing features, \textsc{Generalize} overly specific features, and \textsc{Mask} irrelevant features. The robot then plans under the corrected feature set to optimize a reward function through demonstrations (blue), yielding trajectories that are better aligned with the human's true reward (purple, bottom-right).}
    \label{fig:pipeline}
\end{figure}

\begin{abstract}
Trajectory demonstrations show robots how to behave but not why, leaving them to guess which \textit{features} of the task actually matter. Incorrect guesses produce misaligned internal representations and lead to incorrect robot behaviors. Because task features are often semantically defined, we argue that natural language is better suited for specifying features rather than relying on demonstrations. Although existing representation editing methods have proposed using language to specify new features, only allowing additive edits produces feature sets that are often over-specified or too narrowly scoped, especially in sequential editing tasks. We propose the Flexible Feature Editing Pipeline (\ours), which uses language models to iteratively refine a robot's internal representation of task features from natural language corrections via three complementary operations: \textsc{Add} missing features, \textsc{Generalize} existing specific features, and \textsc{Mask} irrelevant features. We evaluate \ours on 2D navigation scenarios, 7DoF Franka tabletop manipulation tasks, and a real-world study using naturalistic human language corrections and real-robot demonstrations. Comparisons against single-operation and unscaffolded baselines show that all three operations are necessary in interactive scenarios, and that \ours achieves consistently higher normalized reward compared to other feature editing approaches while maintaining compact, transferable representations. Project available at \href{https://mit-clear-lab.github.io/FLEX/}{https://mit-clear-lab.github.io/FLEX/}.
\end{abstract}



\section{Introduction}
\label{sec:introduction}

Imagine a scenario like in Fig.~\ref{fig:pipeline}, where a human asks their personal robot to bring them coffee. While the instruction seems simple, executing it well depends on the \textit{features}, or measurable properties of the environment, that the robot internally uses to represent the task. For example, a user may prefer that the robot stays away from sensitive electronics or that it stays close to the table to avoid damage from spills. Unfortunately, human instructions rarely describe these features explicitly when specifying tasks, forcing the robot to guess which features matter. When the robot guesses incorrectly, the human and robot end up with \textit{misaligned} representations, and the robot fails to complete the task in the way that the human desires.

A common approach to address the representation misalignment problem is to iterate between specifying features and optimizing reward, often through demonstrations~\cite{peng2024adaptive,levine2010feature}. However, demonstrations carry little semantic information, so the robot can only \textit{implicitly} recover which features actually matter to the user. In contrast, natural language contains rich semantic content. When a user says, ``Don't spill coffee on my laptop!'' their language communicates both what matters about the task (avoiding the laptop) and why that matters (preventing damage). Language contains the semantic content that demonstrations lack. We therefore propose that feature specification is better addressed through language, while demonstrations are better-suited for reward optimization once the right features are known.

Even a simple language correction contains a wealth of information when interpreted in context. However, existing methods fail to leverage this fully, either failing to update the robot's underlying representation in a generalizable way~\cite{ren2023knowno} or continually adding features until representations become redundant and inefficient~\cite{peng2024adaptive}. None of these methods account for the full space of ways a feature representation can be wrong: it may be \textit{incomplete} (missing relevant features), \textit{over-specified} (tracking irrelevant features), or \textit{incorrectly abstracted} (tracking a feature at the wrong level of abstraction).

We propose \ours, a flexible feature editing pipeline that uses Language Models (LMs) to adapt a robot's internal representation of task features in response to natural language corrections. As shown in Fig.~\ref{fig:pipeline}, when the human says ``Don't spill coffee on my laptop!'', \ours identifies the correction's intent and applies three operations: \textsc{Add} introduces the laptop as a relevant obstacle, \textsc{Generalize} broadens the newly added feature to all electronics, and \textsc{Mask} removes the now-irrelevant bowl feature. This yields a feature set aligned with the human's true intent. Unlike prior methods that support only a single operation, \ours selectively applies whichever combination of edits the user's language correction warrants. 





\section{Related Work}
\label{sec:related_work}

\noindent\textbf{Reward Learning from Human Feedback.}
Learning from human feedback is necessary to adapt robot policies to new users and tasks~\cite{hadfield2017inverse,christiano2017deep,sadigh2017pref,baraka2026human}. Past work has proposed many feedback signals---such as demonstrations~\cite{chen2025tool,xu2025dexumi, argall2009survey,tang2025mimicfunc}, corrections~\cite{li2021learning,bajcsy2017learning,bobu2018learning}, rankings~\cite{dennler2026improving,myers2022learning,lu2022preference}, and other user behaviors~\cite{liang2024visarl,jeon2020reward,dennler2025contrastive,sheidlower2022keeping}---to adapt robot behavior to user preferences. However, many of the most common forms of feedback are underspecified; for example, similar demonstrations can be consistent with many possible reward functions~\cite{bobu2024alignment,casper2023open,bobu2020quantifying}. Resolving this ambiguity requires large amounts of data that are impractical to collect from non-expert users~\cite{hirota2025active,dennler2025contrastive}. While past work has addressed the data efficiency problem by combining multiple types of feedback~\cite{liang2024visarl,cui2023no,hwang2026language,nader2025quicklap}, these approaches collect data and train rewards offline. We instead focus on robots that adapt at deployment time through iterative user feedback.

\noindent\textbf{Specifying Trajectory Features.}
To make online reward learning tractable for the amount of data that can be collected at deployment time, the robot must restrict its set of reward hypotheses to a low-dimensional space, rather than attempting to learn directly from high-dimensional observations~\cite{bartlett2002rademacher,bobu2022inducing}. Prior work defines this space by using expressive trajectory features that are either designed by experts~\cite{abbeel2004apprenticeship, ziebart2008maximum,tucker2020preference,koyama2020sequential}, learned from data~\cite{levine2010feature, bobu2023sirl,dennler2025contrastive, lee2023user,bobu2021ferl,forsey2026context}, or generated with LMs~\cite{peng2024adaptive, ma2024eureka}. 
The problem of \textit{representation misalignment}~\cite{bobu2024alignment,sucholutsky2025gettingaligned} arises when these feature sets do not capture what the user actually cares about, which results in the robot failing to complete the user's desired task. 
This problem motivates recent works that allow users to directly edit the feature set to align user and robot 
representations~\cite{peng2024adaptive,koyama2020sequential}. 
However, these prior methods support only a single type of edit: adding features~\cite{peng2024adaptive}, removing features~\cite{peng2024learning}, or generalizing the scope of features~\cite{amin2026gift}.
Our work unifies these separate editing operations into a single framework, and demonstrates that all three edit types are necessary to address the full range of representational misalignment types.

\noindent\textbf{Natural Language for Robot Learning.}
A growing body of work incorporates natural language into robot learning pipelines, using language to clarify task goals~\cite{ren2023knowno,zitkovich2023rt,singh2022progprompt,black2025pi_,yin2025gc,wu2025foresight}, redirect ongoing behavior~\cite{shi2024language,cui2023no,liu2023interactive}, or update internal representations such as safety constraints~\cite{abugurain2024language,santos2025updating}. 
However, these approaches treat language primarily as a signal for \textit{how} to perform a task, rather than \textit{which} features of the environment matter~\cite{hwang2026language}. 
We identify that language carries unique semantic information that makes it particularly well-suited for feature editing. LMs are equipped to leverage this data because they are trained on human-generated data and have reasonable priors over shared preferences~\cite{bowen2023humanmodels,amin2026gift,park2023generative}, which allows them to efficiently leverage sparse natural language corrections~\cite{nader2025quicklap,zha2024distilling}. 
The iterative nature of our approach provides advantages over approaches that use language solely to specify task goals from single instances of feedback.

\section{Problem Definition}
\label{sec:problem}
We focus on the problem of learning reward functions that reflect users' underlying preferences, where our goal is to generate robot behavior from limited amounts of user feedback. In this work, the user provides feedback as demonstrations and language.

\textbf{Preliminaries.} We consider the robot to be operating in an MDP defined by $\langle \mathcal{S}, \mathcal{A}, \mathcal{T}, \mathcal{R} \rangle$, with state space $\mathcal{S} \subseteq \mathbb{R}^{d_s}$, action space $\mathcal{A} \subseteq \mathbb{R}^{d_a}$, transition function $\mathcal{T} : \mathcal{S} \times \mathcal{A} \times \mathcal{S} \rightarrow [0,1]$, and reward function $\mathcal{R} : \mathcal{S} \times \mathcal{A} \rightarrow \mathbb{R}$. A solution to this MDP is a policy $\pi : \mathcal{S} \rightarrow \mathcal{A}$ that when rolled out produces a trajectory, $\tau = (s^0, a^0, s^1, a^1,\dots, s^T)$. Following past work in inverse reinforcement learning~\cite{ng2000algorithms,abbeel2004apprenticeship}, we model the reward as a linear combination of state features, $\Phi(s)$, such that the reward over a full trajectory is expressed as:
\begin{equation}
    R_\theta(\tau) = \sum_{i=0}^T \theta^\top \Phi(s^i) = \theta^\top \Phi(\tau)\enspace.
\end{equation}
Here, $\theta$ represents the user's preferences as a vector of coefficients that express a weighted tradeoff across the task features, $\Phi$.

\textbf{Reward Optimization.}
In practice, the robot doesn't know the reward function $R_\theta$ and must learn it from human feedback. This problem is well-studied in domains where the correct $\Phi$ is known \textit{a priori}~\cite{abbeel2004apprenticeship,ziebart2008maximum,sadigh2017pref,bajcsy2017learning}. One standard approach is to estimate the preference weights, $\theta$, by collecting samples of the user's preferred behavior as a dataset of demonstrations, $\mathcal{D} = \{\tau_i\}_{i=0}^N$.

While our framework is agnostic to the specific reward learning algorithm used, we adopt the maximum entropy principle~\cite{jaynes1957information,ziebart2008maximum,ng2000algorithms}. This principle models the human as a \textit{noisily rational} agent that intends to provide optimal trajectories but does not always execute them perfectly, as captured by the Boltzmann distribution over trajectories:
\begin{equation}
    p(\tau \mid \theta) \propto \exp(\theta^\top \Phi(\tau))\enspace.
\end{equation}
The Boltzmann distribution assigns higher probability to trajectories with higher reward, while still allowing for suboptimal behavior. The likelihood of the full demonstration dataset is then $p(\mathcal{D} \mid \theta) = \prod_{\tau \in \mathcal{D}} p(\tau \mid \theta)$, and the robot can estimate $\theta$ via Bayes' Rule:
\begin{equation}
    p(\theta \mid \mathcal{D})\propto p(\mathcal{D} \mid \theta)p(\theta)\enspace,
\end{equation}
where $p(\theta)$ is a uniform prior over $\theta$. Once the robot infers $\theta$, it can solve $\tau^*= \arg\max_{\tau} \;\theta^\top\Phi(\tau)$ to recover the user's preferred trajectory.

%
\noindent\textbf{Feature Specification.}
While this approach is effective when the right features are available, 
it assumes that the relevant features are specified before deployment. 
This assumption is often unrealistic in unstructured and evolving human-centered environments. 
In this work, we consider the more general problem of jointly estimating the representation $\Phi$ alongside the reward weights $\theta$~\cite{peng2024adaptive}:
\begin{equation}
    \theta^*, \Phi^* = \arg\max_{\theta, \Phi} \; p(\Phi, \theta \mid \mathcal{D})\enspace.
\end{equation}
%
Prior works have addressed this problem by decomposing the problem into two iterative steps: \textit{feature specification}, 
which refines the feature set $\Phi^{k+1}$ under a fixed $\theta^k$, and \textit{reward optimization}, which estimates $\theta^{k+1}$ using a fixed $\Phi^{k+1}$ at each iteration~\cite{levine2010feature,peng2024adaptive}.

The goal of feature specification is to use data collected from the user to estimate a refined feature set at each iteration, given the current feature set $\Phi^k$ and reward weights $\theta^k$:
\begin{equation}
    \Phi^{k+1} = \arg\max_\Phi \; p(\Phi \mid \Phi^k, \mathcal{D}; \theta^k)\enspace.
    \label{eq:rep_learning_demos}
\end{equation}
Past work uses demonstrations for both steps of the process~\cite{levine2010feature,peng2024adaptive}, but we argue this is suboptimal: a single demonstrated trajectory may be consistent with many possible representations. 
While one solution is to collect more demonstrations, 
this is labor-intensive and places a heavy burden on a signal that does not efficiently carry representational information.

Language corrections, by contrast, can express representational content directly, making them a more natural and efficient signal for feature specification.
Our key insight is that feature specification can better leverage \textit{semantic} information such as language corrections, whereas demonstrations are better for communicating \textit{preferential} information for reward optimization.
We thus extend the feature specification problem to leverage iterative language corrections, $\ell^k$:
\begin{equation}
    \Phi^{k+1} = \arg\max_\Phi \; p(\Phi \mid \Phi^k, \ell^k; \theta^k)\enspace.
    \label{eq:rep_learning_language}
\end{equation}
%
We next describe our algorithm for performing feature specification with language feedback.


\section{Flexible Feature Editing Pipeline}
\label{sec:methods}
The Flexible Feature Editing Pipeline (\ours) uses natural language to specify features by interpreting semantic feedback through a sequence of complementary operations.

At each iteration $k$, the robot executes its current best trajectory $\tau^k = \arg\max_\tau \theta^{k\top}\Phi^k(\tau)$, which reflects both its current representation $\Phi^k$ and weights $\theta^k$. The human's desired behavior is an imagined trajectory $\tau^* = \arg\max_\tau \theta^{*\top}\Phi^*(\tau)$ that is optimal under $\Phi^*$ and reward weights $\theta^*$. We assume that the human observes $\tau^k$, identifies the misalignment between $\tau^k$ and $\tau^*$, and provides a language correction $\ell^k$ to communicate this misalignment: $\ell^k \sim P(\ell \mid \tau^*, \tau^k)$.

Since $\tau^*$ and $\tau^k$ are deterministic functions of
$(\Phi^*, \theta^*)$ and $(\Phi^k, \theta^k)$ respectively, 
we can write $\ell^k \sim P(\ell \mid \Phi^*, \theta^*, \Phi^k, \theta^k)$. In this paper, 
we use $\ell^k$ to infer which features matter ($\Phi^*$), but not to infer how much they matter ($\theta^*$). Under this assumption, $\ell$ is conditionally independent of $\theta^*$ and $\theta^k$ given $\Phi^*$ and $\Phi^k$, simplifying to $\ell^k \sim P(\ell \mid \Phi^*, \Phi^k)$. 

This gives us a probabilistic model of human language corrections.
Since we are interested in estimating the human's true representation $\Phi^*$, we apply Bayes' rule to obtain:
\begin{equation}
    P(\Phi^* \mid \Phi^k, \ell^k) \propto P(\ell^k \mid \Phi^*, \Phi^k) \cdot P(\Phi^*\mid\Phi^k)\enspace,
    \label{eq:phi_given_language}
\end{equation}
where we assume a uniform prior $P(\Phi^* \mid \Phi^k)$, reflecting the assumption that the human's true representation $\Phi^*$ is independent of what the robot currently knows $\Phi^k$. This exactly corresponds to Eq.~\ref{eq:rep_learning_language} with $\theta^k$ dropped under our representational feedback assumption. 
Rather than computing the posterior in Eq.~\ref{eq:phi_given_language} explicitly, we approximate
$\Phi^{k+1} = \arg\max_{\Phi} P(\Phi \mid \Phi^k, \ell^k)$ directly using a language model, which is well suited to proposing
feature sets that explain the observed correction $\ell^k$ given the
current feature set $\Phi^k$: $\Phi^{k+1} \approx f_{\text{LM}}(\Phi^k, \ell^k)$.
%
%
%

We break down $f_\text{LM}(\featuresetcurr, \correctionk)$ into a set of three distinct editing operations in the following order: \textsc{Add}, \textsc{Generalize}, and \textsc{Mask}. 
While prior work has proposed individual operations for feature editing, we show that a single language correction can simultaneously motivate several distinct types of edits, and that all three are necessary for robust reward learning.
Each operation takes as input a textual description of the task and environment with the user's natural language correction and maps the current feature set to a refined one, $\Phi^k \rightarrow \Phi^{k+1}$.

\noindent\textbf{The \textsc{Add} Operation.} The \textsc{Add} operation takes the user's correction \correction and the current feature set \featureset as input to expand the feature set to consider new objectives: $\textsc{Add}( \correction, \featureset) = \Phi'$, where $\Phi' \supseteq \Phi$. 

\textsc{Add} queries an LM twice: once for the name of the new feature and once to generate code that implements the function that computes this new feature. First, the LM takes \correctionk and \featuresetcurr as input to jointly (1) estimate a boolean \textit{sufficiency score}, $b \in \{0,1\}$, that determines if the existing feature set \featuresetcurr captures the content of the language correction, and (2) compute a \textit{proposed feature} $\phi'$ that would capture the content of the language correction. The LM returns the proposed feature if \featuresetcurr did not explain the user's correction or the null set if \featuresetcurr was sufficient:
\begin{equation}
    f_\text{LM}(\featuresetcurr, \correctionk) = 
    \begin{cases} 
    \phi' & \text{if } b=0 \\
    \emptyset & \text{otherwise}
    \end{cases}
\end{equation}
If a new feature $\phi'$ is returned, we append it to the existing set: $\Phi^{k+1} = \Phi^k \cup \phi'\enspace.$
%
%

\noindent\textbf{The \textsc{Generalize} Operation.} The \textsc{Generalize} operation takes the user's correction \correction, and the current feature set \featureset as input, and returns an augmented feature set: $\textsc{Generalize}(\correction, \featureset) = \Phi'$, where  $|\Phi'| = |\Phi|$. 

The augmented feature set contains features that are more general than the input feature set. For example, if a user tells a robot carrying water to stay away from a laptop, the higher-level intent may be to ``stay away from electronic devices''. The resulting feature generalization produces a new feature that not only indicates distance to laptops, but also distance to other electronic devices, such as cell phones or calculators. We leverage the common-sense reasoning capabilities of LMs~\cite{amin2026gift} to explicitly extract a higher-level intent $\omega$ behind the correction through zero-shot prompting:
\begin{equation}
    f_{LM}(\featuresetcurr, \correctionk) = \omega\enspace.
\end{equation}
For each feature $\phi_i \in \Phi^k$, we determine if that feature can be generalized to new objects that match the high-level intent, $\omega$. This process produces a new feature based on the high-level intent:
\begin{equation}
    f_{LM}(\correctionk, \phi_i, \omega) = \phi_i^{gen}\enspace.
\end{equation}
In practice, we create these generalized features by assuming objects have a set of text-based descriptive properties to generalize across. 
We implement the generalized feature with code generated by the LM, which replaces its original implementation, $\Phi^{k+1} = (\Phi^k \backslash \{\phi_i\}) \cup \{\phi_i^{gen}\}  \enspace. $
%
%

\noindent\textbf{The \textsc{Mask} Operation.} The \textsc{Mask} operation takes the user's correction \correction and the current feature set \featureset as input to reduce the current feature set to remove extraneous or irrelevant features: $\textsc{Mask}(\correction, \featureset) = \Phi'$, where $\Phi' \subseteq \Phi$.

To compute the \textsc{Mask} operation, we use an LM to generate a binary mask over the feature set, $\Phi$. For each feature, the LM estimates a boolean relevancy score, $m_i\in \{0,1\}$. We query the LM once per feature to determine whether each $\phi_i \in \Phi^k$ remains relevant given the current context:
\begin{equation}
    m_i = f_\text{LM}(\featuresetcurr, \correctionk) = 
    \begin{cases} 
    1 & \text{if relevant}(\phi_i, \correctionk) \\
    0 & \text{otherwise}
    \end{cases} \enspace.
\end{equation}

\noindent\textbf{Operation Order.} \ours applies operations in the order \textsc{Add} $\rightarrow$ \textsc{Generalize} $\rightarrow$ \textsc{Mask}, reflecting dependencies between the operations. \textsc{Add} runs first to ensure newly added features are fully processed. \textsc{Generalize} follows, and \textsc{Mask} runs last so that masking decisions are made on the features' final forms. Other orderings require re-iteration to reach a consistent representation.

\begin{figure}[t]
  \centering
    \includegraphics[width=0.9\linewidth]{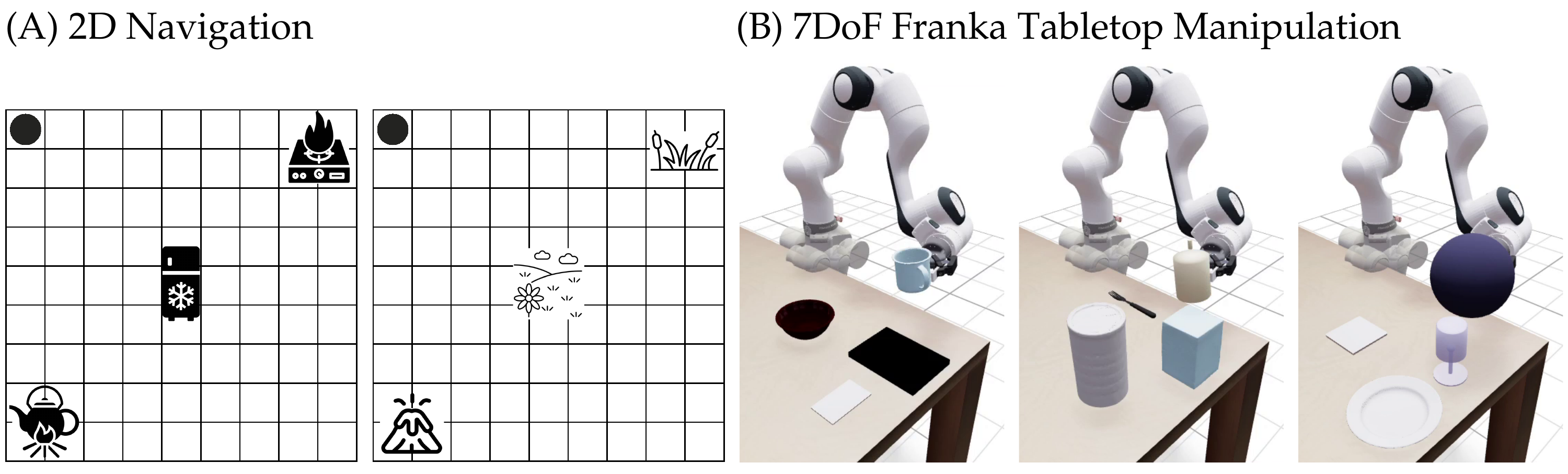}
  \caption{Simulation domains. (A) A 9x9 2D navigation environment, and (B) 7DoF Franka, a tabletop manipulation environment. Each domain includes object variants to test generalization.}
  \label{fig:sim_env}
  \vspace{-1.5em}
\end{figure}

\section{Experiments}
\label{sec:exp}
We evaluate our algorithm in simulated and real environments through three key research questions that progressively relax assumptions about robot and human behavior: (1) Are all three operations necessary for aligning human and robot representations? (2) Does \ours perform well in continuous control settings with sequential corrections? (3) Does \ours perform well with the natural and idiosyncratic feedback that real humans provide?

\noindent\textbf{Domains.} Our experiments span two domains: \textit{2D Navigation} and \textit{7DoF Franka Tabletop Manipulation}, shown in Fig. \ref{fig:sim_env}.
In the \textit{2D Navigation} domain, each scenario occurs in a 9-by-9 2D grid where a robot moves from a start position to a goal position while avoiding obstacles. 
In the \textit{7DoF Franka Tabletop Manipulation} domain, a robot arm manipulates various objects in a tabletop setting by moving from a start configuration to a goal configuration while avoiding objects. 

\noindent\textbf{Baselines.} We evaluate \ours against two single-operation baselines and one unscaffolded LM baseline. (1) \textbf{\ours} uses all operations described in Section~\ref{sec:methods}: \textsc{Add}, \textsc{Generalize}, and \textsc{Mask}. (2) \textbf{ALGAE}~\cite{peng2024adaptive} is a baseline that only adds new features based on the human's language correction and demonstrations. (3) Mask-From-Exhaustive (\textbf{MFE}) starts with an exhaustive list of features and queries an LM with the human's language correction to mask out all irrelevant features. (4) Unscaffolded (\textbf{USF}) queries the LM once to directly generate the full updated $\Phi^{k+1}$ from the correction, demonstration, and current $\Phi^k$ without decomposing the edit into our three operations. All experiments use GPT-4o as the LM.

\noindent\textbf{Metrics.} All robot scenarios involve a robot moving from a starting configuration to a goal configuration with various objects in the environment. In all scenarios, we randomize start, goal, and object positions across five seeds for evaluation. Our main evaluation metric is \textit{normalized reward}, which is computed as $\hat{R} = (R - R_{\text{worst}}) / (R_{\text{best}} - R_{\text{worst}})$. $R_{\text{best}}$ and $R_{\text{worst}}$ are the rewards of the ground-truth optimal and worst-case trajectories, respectively. A score of $\hat{R} = 1$ indicates optimal behavior and a score of $\hat{R} = 0$ indicates worst-case behavior.

\subsection{Isolation of \ours's Components (RQ1)}

Our first research question evaluates whether all three operations of \ours are necessary.

\noindent\textbf{Experimental Details.} We evaluate RQ1 using two variants of the 2D Navigation domain to demonstrate \ours's usefulness across different semantic contexts. In KitchenWorld (Fig.~\ref{fig:sim_env}A, left), the robot must avoid hot appliances (a stove and a kettle) with a fridge as a \textit{distractor object}---an object that is present in the environment but does not affect the human's preference for how the robot completes the task. In LavaLand (Fig.~\ref{fig:sim_env}A, right), the robot must avoid dangerous terrain (a swamp and lava) with a meadow as a distractor terrain.

Across these two variants, we define seven distinct scenarios that each test a combination of \ours's three operations. Each algorithm we test takes a misaligned feature set and a natural language correction as input, and returns an updated feature set. We use the resulting updated feature set and the simulated expert demonstrations to learn a reward. To evaluate how well each of these algorithms performs, we compare our estimated reward to a ground-truth reward that is pre-defined for each scenario using the normalized reward metric described above. We provide additional details about the specific configuration of each scenario in the Appendix.

\begin{table}[t!]
  \setlength{\belowcaptionskip}{-6pt}
  \centering
  \resizebox{\textwidth}{!}{%
  \begin{tabular}{lccccccc}
    \toprule
    \textbf{Method} & \textbf{Add Only} & \textbf{Gen. Only} & \textbf{Mask Only} & \textbf{Add+Gen.} & \textbf{Add+Mask} & \textbf{Mask+Gen.} & \textbf{All Three} \\
    \midrule
    \textsc{ALGAE}
      & $\mathtt{\mathbf{1.00 \pm .00}}$
      & $\mathtt{.59 \pm .02}$
      & $\mathtt{.95 \pm .03}$
      & $\mathtt{.77 \pm .02}$
      & $\mathtt{.59 \pm .02}$
      & $\mathtt{.63 \pm .02}$
      & $\mathtt{.69 \pm .03}$ \\
    \textsc{MFE}
      & $\mathtt{.95 \pm .01}$
      & $\mathtt{.70 \pm .02}$
      & $\mathtt{\mathbf{1.00 \pm .00}}$
      & $\mathtt{.75 \pm .02}$
      & $\mathtt{.43 \pm .01}$
      & $\mathtt{.67 \pm .02}$
      & $\mathtt{.83 \pm .02}$ \\
    \textsc{USF}
      & $\mathtt{.64 \pm .06}$
      & $\mathtt{.72 \pm .03}$
      & $\mathtt{.90 \pm .02}$
      & $\mathtt{.80 \pm .01}$
      & $\mathtt{.99 \pm .01}$
      & $\mathtt{.78 \pm .04}$
      & $\mathtt{.78 \pm .03}$ \\
    \midrule
    \textbf{\ours (Ours)}
      & $\mathtt{\mathbf{1.00 \pm .00}}$
      & $\mathtt{\mathbf{.91 \pm .03}}$
      & $\mathtt{\mathbf{1.00 \pm .00}}$
      & $\mathtt{\mathbf{1.00 \pm .00}}$
      & $\mathtt{\mathbf{1.00 \pm .00}}$
      & $\mathtt{\mathbf{1.00 \pm .01}}$
      & $\mathtt{\mathbf{.99 \pm .01}}$ \\
    \bottomrule
    \end{tabular}%
  }

  \caption{Average normalized reward $\hat{R}$ (mean\,$\pm$\,SE) across seven 2D Navigation scenarios. \textbf{Bold} indicates the best-performing method. \ours maintains strong performance across all scenarios.\label{tab:grid_results}}
  \vspace{-1em}
\end{table}

\noindent\textbf{Results.} We show the results of the average normalized reward for each scenario in Table~\ref{tab:grid_results}. \textbf{\ours} achieves the highest normalized reward across all seven misalignment scenarios. The single-operation baselines perform well on scenarios that use the edit types they are best-suited for: \textbf{ALGAE} performs competitively for the \textsc{Add}-only scenario and \textbf{MFE} performs competitively for the \textsc{Mask}-only scenario. However, their performance substantially degrades when the scenarios require other operations, particularly \textsc{Generalize}. \textbf{USF} is highly inconsistent, displaying a high range of mean rewards. Without operation-level scaffolding, its success on a given operation is difficult to anticipate, achieving only 0.64 reward in the simplest \textsc{Add Only} scenario yet 0.99 reward when combining \textsc{Add+Mask}. Together, these results show that all three operations are necessary to address a range of feature misalignments, and that scaffolding them explicitly makes \ours more robust at learning users' preferences than either restricted or unstructured editing.

\subsection{Evaluation in Simulated Continuous Control Domain (RQ2)}

Our second research question examines sequential feature edits in a continuous control task.

\begin{figure}[h!]
    \centering
    \includegraphics[width=\textwidth]{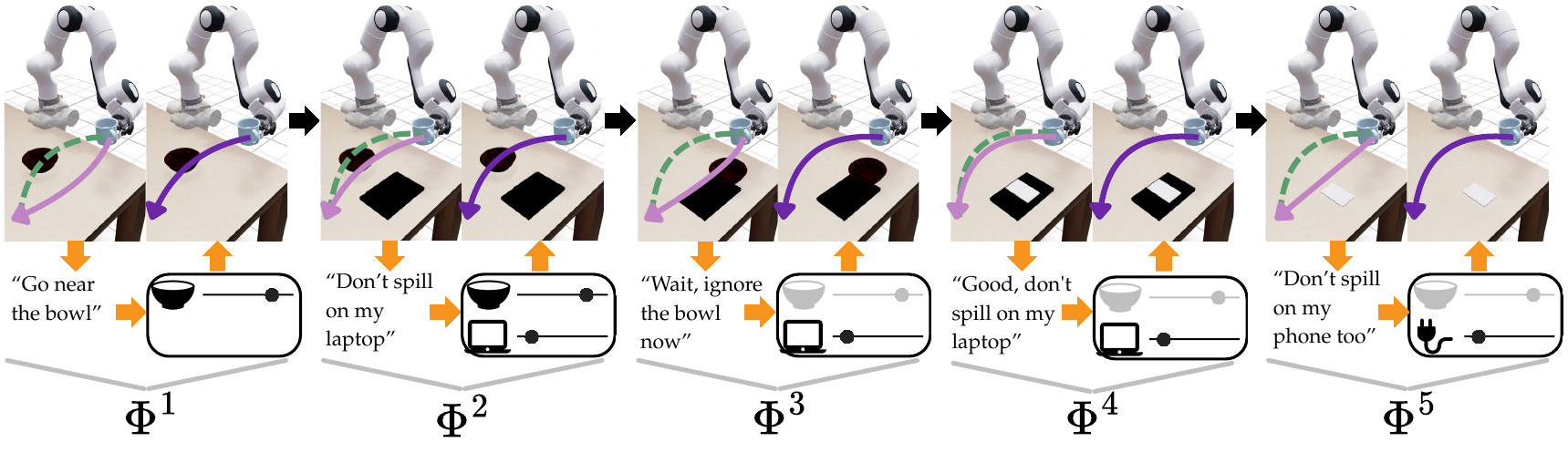}
    \caption{Five-stage Franka correction sequence. At each stage, the user observes a trajectory (lavender) that is misaligned with their ground truth trajectory (green). They then provide a language correction, prompting the robot to update its internal representation $\Phi^k$. Although corrections may immediately realign behavior (purple), environment changes in the subsequent stages can expose new misalignments. Stages are designed so that all three \ours operations become necessary at some point during the sequence, reflecting realistic long-term deployment conditions. \label{fig:franka_seq}
    }
    \vspace{-1em}
\end{figure}

\noindent\textbf{Domain Details.} RQ2 evaluates \ours in a task where a person performs a series of edits, reflective of the intended use of \ours. RQ2 considers three variants of the 7DoF Franka domain: CoffeeRobot, CandleRobot, and BallRobot. In CoffeeRobot (Fig.~\ref{fig:sim_env}B, left), the robot must carry a full cup of coffee while avoiding electronic devices (a laptop and a phone) with a wooden bowl as a distractor object. In CandleRobot (Fig.~\ref{fig:sim_env}B, middle), the robot must carry a lit candle while avoiding flammable objects (a box of tissues and a paper towel roll) with a fork as a distractor object. In BallRobot (Fig.~\ref{fig:sim_env}B, right), the robot must carry a bowling ball while avoiding fragile objects (a wineglass and a ceramic plate) with a napkin as a distractor object. Rather than evaluating each correction in isolation, we construct five-stage correction sequences (Fig.~\ref{fig:franka_seq} shows the CoffeeRobot sequence) in which edits stack, with each stage starting from the representation $\Phi^k$ produced by the previous stage. Between stages, we change the objects in the environment to expose new misalignments and to show how all three \ours operations can become necessary over long-term deployment.

\begin{figure}[h!]
    \centering
    \includegraphics[width=\textwidth]{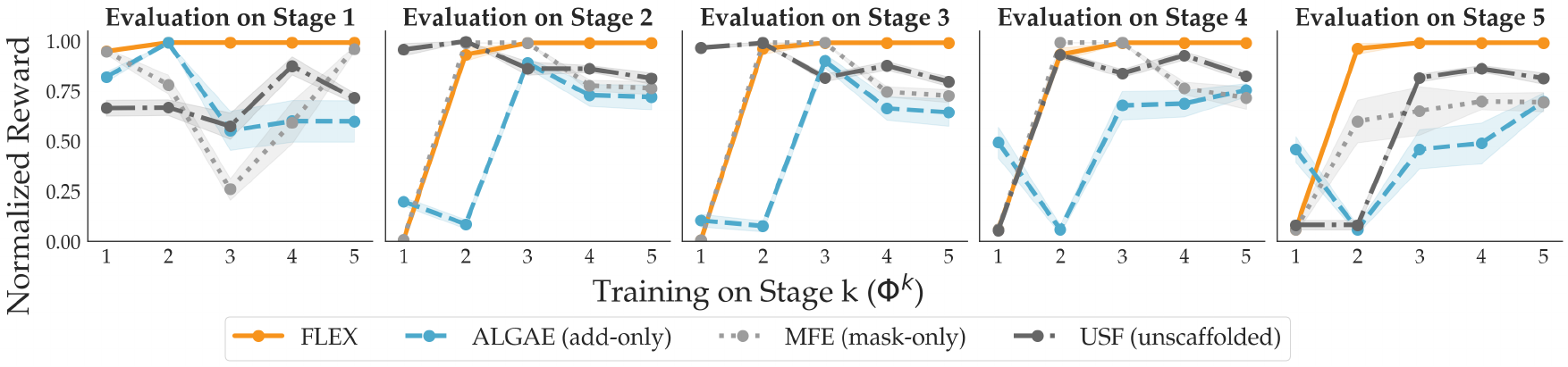}
    
    \caption{
    Average normalized reward $\hat{R}$ across five stages of the Franka correction sequences, averaged across task variants. The shaded regions indicate $\pm 1$ SE. The x-axis denotes the iteration of $\Phi^k$, the feature set learned after training on Stage $k$. \ours sustains high performance across all stages, while all baseline performances degrade as training stages accumulate.}
    \label{fig:franka_results}
    \vspace{-1.5em}
\end{figure}

\noindent\textbf{Results.} 
Each panel in Fig.~\ref{fig:franka_results}, ``Evaluation on Stage $j$,'' shows the average normalized reward earned on stage $j$ by the representation $\Phi^k$ learned after training on stage $k$. A good representation should perform well on the stage it was trained on, on earlier stages (no forgetting), and on later stages (generalization). We find that \ours learns the user's ground-truth reward function after one or two corrections and maintains high performance across all evaluation stages as corrections accumulate. Baseline methods instead require additional corrections to learn the user's reward, and their performance degrades as corrections accumulate because they gather spurious features, fail to re-abstract earlier features, and overfit to the stage they were trained on.

\subsection{Evaluation with Real Corrections and Demonstrations (RQ3)}

Our third research question explores realistic human language and demonstrations.

\noindent\textbf{User Study Setup.} We conducted an in-person, within-subjects study with $N=12$ participants recruited from a university pool (6 women, 6 men). Five participants were novice robot users and self-reported little to no prior familiarity with robots on a 7-point scale. Each participant interacted with three methods (\textbf{\ours}, \textbf{ALGAE}, \textbf{MFE}). Each method was randomly paired with one of the three Franka task scenarios from Fig.~\ref{fig:sim_env}B. We randomized and counterbalanced the order of method and task presentation to control for ordering effects. This study was IRB-approved, and we compensated participants \$25 for a one-hour session.

Before the experiment, participants filled out a consent form and demographics questionnaire. Participants then spent $\sim$10 minutes performing a tutorial that familiarized them with the interface’s button and mouse controls (shown in App.~\ref{app:likert}). In the tutorial, each participant completed two practice corrections on a task that was different from the study scenarios.

During the experiment, the user progressed through three experimental blocks. Each block consisted of five sequential correction stages (example in Fig.~\ref{fig:franka_seq}). In each stage, participants observed the robot’s current trajectory alongside a goal trajectory optimized for the ground-truth reward. To keep conditions consistent across participants, each task had a pre-defined ground-truth preference (e.g., avoiding electronics when carrying coffee), and participants gave typed natural language corrections to correct the robot’s behavior so that it matched the pre-defined preference. The robot’s learned reward updated in real-time after each correction, so the ``current’’ trajectory shown in subsequent rounds reflected the participant’s full correction history. After each five-stage block, participants completed a Likert questionnaire to evaluate how well the algorithm performed (items in App.~\ref{app:likert}).

\begin{figure}[h!]
    \centering
    \includegraphics[width=0.8\textwidth]{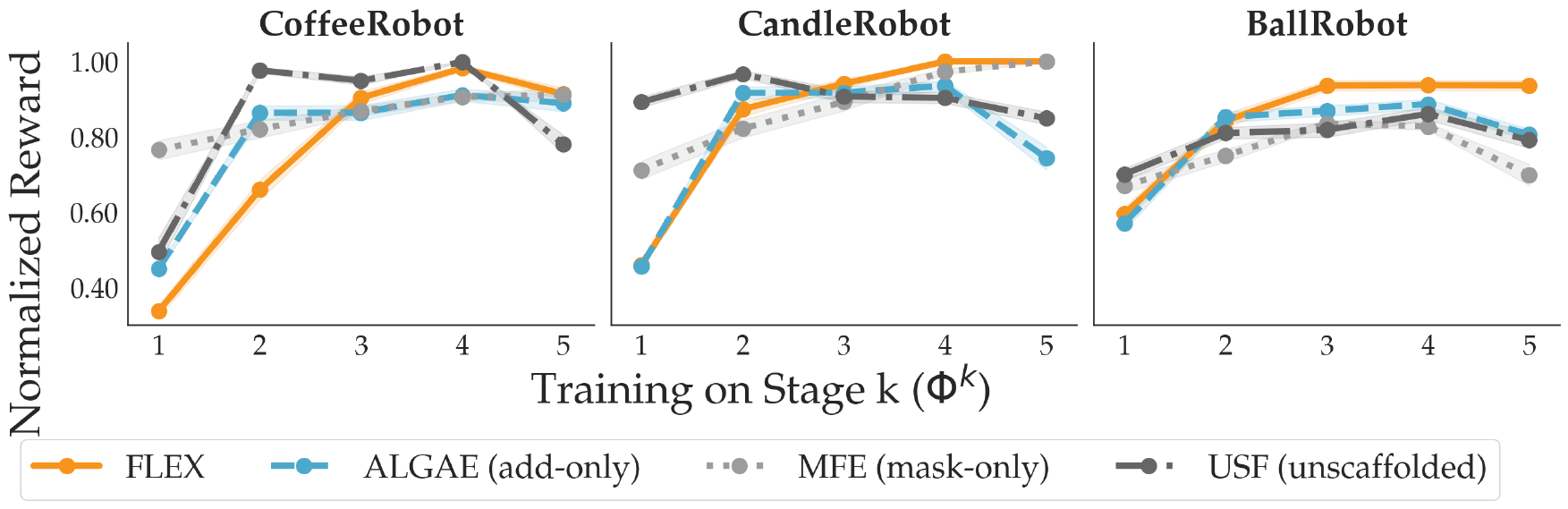}
    
    \caption{
    Average normalized reward $\hat{R}$ of representations learned from real human language corrections and demonstrations across the three variants of 7DoF Franka correction sequences. The shaded regions indicate $\pm 1$ SE. The x-axis denotes the iteration of $\Phi^k$, the feature set learned after training on Stage $k$. \ours sustains its high performance after learning the user's reward, while all baseline performances degrade as their feature sets accumulate spurious features.}
    \label{fig:real_results}
    \vspace{-1.5em}
\end{figure}

\noindent\textbf{Results.} Participants viewed robot behavior in simulation rather than on the physical robot for safety reasons, since some baselines produced erratic arm motions. To evaluate whether corrections collected this way transfer to real-world settings, we separately ran offline learning using participants’ real corrections with a database of expert-collected demonstrations on the physical Franka, shown in Fig.~\ref{fig:real_results}. We found that \ours successfully learns the user's reward by the end of the sequence. This demonstrates the capability of \ours to adapt to realistic and imprecise user feedback.

\begin{figure}[h!]
    \centering
    \includegraphics[width=0.8\textwidth]{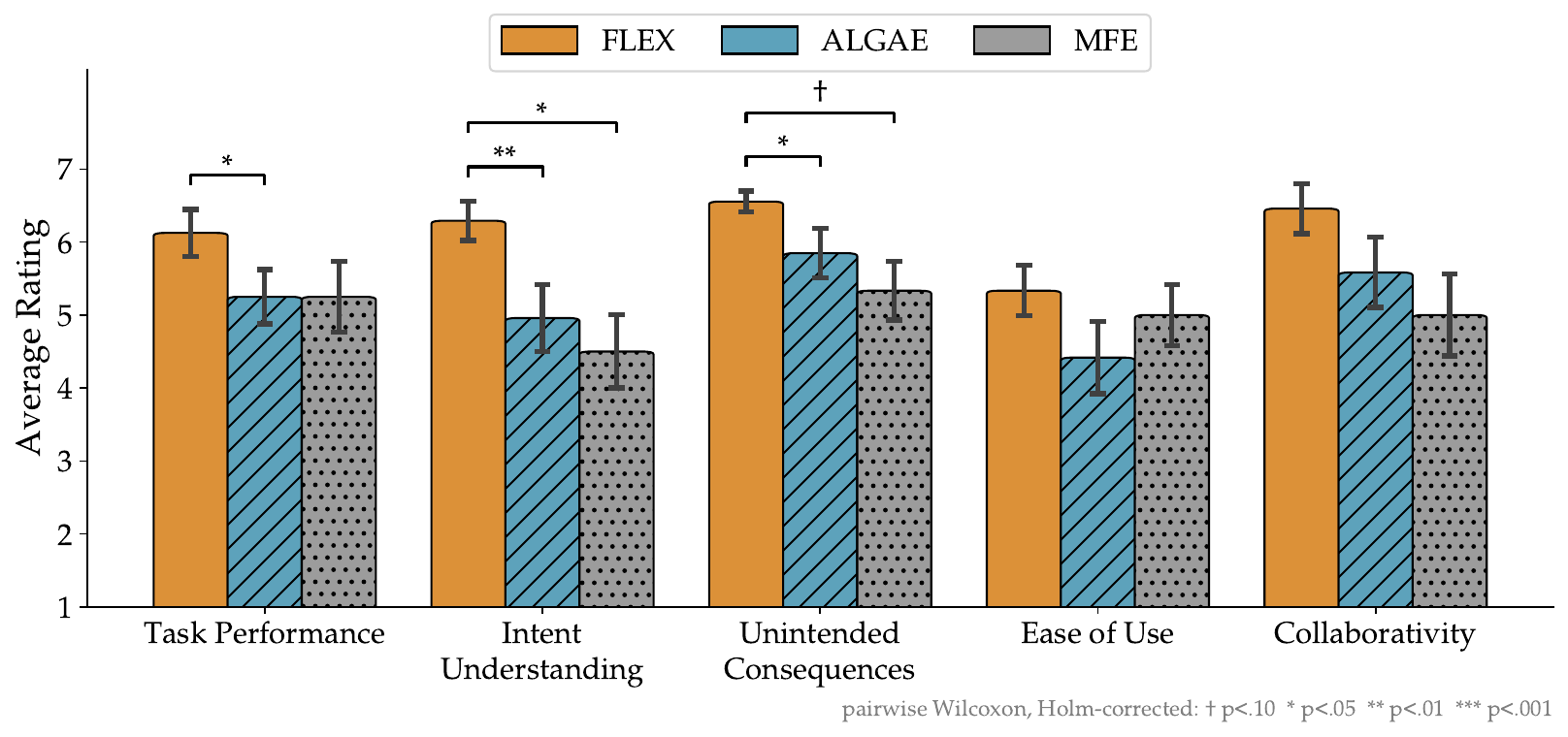}
    \caption{
    Participants' Likert ratings ($N=12$) for \textbf{\ours}, \textbf{ALGAE}, and \textbf{MFE} across five dimensions: task performance, intent understanding, robustness to unintended consequences, ease of use, and collaborativity. Bars show the average rating (1--7 scale) with error bars denoting $\pm 1$ SE. Participants rated \ours highest on all five dimensions. Markers denote significant differences between \ours and each baseline (pairwise Wilcoxon signed-rank tests, Holm-corrected).}
    \label{fig:likert_results}
    \vspace{-1.25em}
\end{figure}

Fig.~\ref{fig:likert_results} shows participants’ Likert responses. Using pairwise Wilcoxon signed-rank tests with Holm-Bonferroni correction (all reported $p$-values are Holm-adjusted), we found that participants rated \ours significantly higher than ALGAE on task performance ($p=.035$), intent understanding ($p=.006$), and robustness to unintended consequences ($p=.047$). \ours significantly outperformed MFE on intent understanding ($p=.031$) and marginally outperformed MFE on robustness to unintended consequences ($p=.074$). \ours also received the highest ratings for ease of use and collaborativity, though the differences were not significant.

\section{Conclusion}
In this work, we introduced \ours, a framework that uses language corrections to perform \textit{feature edits} through three distinct operations: \textsc{Add} new features, \textsc{Generalize} existing features, and \textsc{Mask} irrelevant features. We evaluate \ours across 2D navigation and 7DoF manipulation domains using simulated and real-world user feedback. Across our experiments, we find that \ours accurately learns desired robot behaviors under several types of misspecified feature sets. Furthermore, we demonstrate that \ours transfers effectively from simulated domains to realistic human feedback.


\noindent\textbf{Limitations.} \ours focuses on quasi-static manipulation of rigid objects, where task-relevant features are naturally expressible in language; dynamic tasks or deformable objects may require additional modalities beyond language to specify relevant features. \ours also relies on LM common-sense reasoning, which easily handles widely held preferences but breaks down for personal or counterintuitive ones. For corrections that jointly encode $\Phi$ and $\theta$ (e.g., ``Stay \textit{even further} away from the laptop''), \ours ignores the contribution to $\theta$, as we assume $\ell_k$ is conditionally independent of $\theta^*, \theta_k$. Finally, we present a closed-loop user study conducted in simulation, supported by hardware feasibility checks, rather than closed-loop on a physical robot. Future work could address these limitations by grounding corrections in multimodal observations, developing personalized user models to better support unique users, relaxing the conditional independence assumption to handle language that encodes both $\Phi$ and $\theta$, and conducting a full closed-loop user study on physical hardware.

\clearpage
\acknowledgments{This work has taken place in the Collaborative Learning and Autonomy Research Lab (CLEAR Lab) at MIT and was partly supported by the MIT Generative AI Impact Consortium and the MIT Schwarzman College of Computing’s Social and Ethical Responsibilities of Computing program. Any opinions, findings, and conclusions expressed in this material are those of the authors and do not necessarily reflect the views of the sponsors.}


\bibliography{example}  

\section{Appendix}

\subsection{LM Prompts}
\begin{promptbox}[Add: Generate Feature]
        This is an environment config: \textcolor[HTML]{F7941D}{\{environment\_description\}}\\

        The following is an instruction for a task that the agent needs to complete in this environment. instruction = \textcolor[HTML]{F7941D}{\{instruction\}}\\

        The following is a history of corrections that the agent has been provided. history = \textcolor[HTML]{F7941D}{\{corrections history\}}\\

        The following is the most recent addendum to the original instruction after the agent asked for more clarification about the task. addendum = \textcolor[HTML]{F7941D}{\{correction\}}\\
        
        This is the current list of features that we already know: \textcolor[HTML]{F7941D}{\{$\Phi^k$\}}.\\

        The feature list contains computable numeric values (e.g. "distance from agent to object X") derived from scene state. These features will be used by a downstream model that has NO access to the instruction, corrections, or addendum text -- only to the feature values.\\

        Consider adding a new feature to this list to explain all parts of the instruction and addendum. Carefully consider all parts of the addendum, especially if multiple objects are mentioned.\\

        1. If adding a new feature would not provide any new information that helps explain the instruction or addendum -- for example, if the existing features already capture all relevant objects, properties, or relationships -- do not propose any additional feature and return "NO NEW FEATURE". Do NOT propose a feature merely because an object exists in the environment or because it could be generally relevant in other contexts -- only if the instruction or addendum clearly implies that object or property matters for the task. If the existing features already cover all relevant entities or relationships mentioned or implied by the instruction and addendum, respond exactly with: NO NEW FEATURE\\

        Add at most one sentence of explanation with specific references to the addendum\\

        2. Only propose a new feature if it clearly adds missing information needed to explain the instruction or addendum. You may provide the name of the specific feature in this environment that explains the instruction and addendum. If there are multiple of the same type of object, please give the average distance to all objects of that type.\\
        
        Do not mix multiple types or objects in the feature. Please specify a feature (only one feature e.g. if there are multiple possible explanations choose only one most likely object) that we can write code to compute from the position of the agent to other objects in the scene. Please give a different feature in your answer (e.g. if distance to goal feature is already in the list, do not give distance to goal again). Ground your answer in visible objects in the scene, and give an answer (only 1 specific answer (e.g. distance to heavy objects)) that we can compute relative to the position of the agent.\\
        
        Follow your answer with one newline and at most one sentence of explanation with specific references to the addendum.
\end{promptbox}

\begin{promptbox}[Generalize: Extract Intent]
    This is an environment config: \textcolor[HTML]{F7941D}{\{environment\_description\}}\\

    The following is an instruction for a task that the agent needs to complete in this environment. instruction = \textcolor[HTML]{F7941D}{\{instruction\}}\\

    The following is a history of corrections that the agent has been provided. history = \textcolor[HTML]{F7941D}{\{corrections history\}}\\
    
    The following is the most recent addendum to the original instruction after the agent asked for clarification about the task. addendum = \textcolor[HTML]{F7941D}{\{correction\}}\\

    Extract the intent behind the human's addendum in at most 2 sentences, explaining the purpose or reasoning behind the addendum. Focus on what the agent needs to achieve and why the human may have said this particular addendum, including any important context or conditions that affect the task. Do not base the explanation on specific locations, as this may change in later scenarios.
\end{promptbox}

\begin{promptbox}[Generalize: Generalize Feature Names]
    A \textcolor[HTML]{F7941D}{\{"robot"/"robot arm"\}} is in a \textcolor[HTML]{F7941D}{\{"2D gridworld"/"3D tabletop"\}} environment with the following objects: \textcolor[HTML]{F7941D}{\{environment\_description\}}\\
    
    The following is an instruction for a task that the agent needs to complete in this environment. instruction = \textcolor[HTML]{F7941D}{\{instruction\}}\\

    The following is a history of corrections that the agent has been provided. history = \textcolor[HTML]{F7941D}{\{corrections history\}}\\
    
    The following is an addendum to the original instruction after the agent asked for clarification about the task. addendum = \textcolor[HTML]{F7941D}{\{correction\}}\\

    \textcolor[HTML]{F7941D}{\{intent (from "Generalize: Extract Intent")\}}\\

    The robot has the following set of features: \textcolor[HTML]{F7941D}{\{$\Phi^k$\}}\\

    For each feature, please determine whether the intent is semantically relevant for that feature.\\

    If yes (and only if yes), generalize the feature name using properties of objects in the environment that directly reflect the semantic intent of the instruction or addendum, without creating spurious correlations. \\
    
    Do not generalize based on properties that are not directly connected to the instruction intent. Do not generalize to multiple properties. Feature names may remain unchanged if they already sufficiently capture the semantic intent. Feature names may remain unchanged if they are already sufficiently general. Do not simply rename a feature that points to the same property. Features that share the same semantic relevance should be generalized to the same name. Only generalize using object properties explicitly listed in the environment.\\

    Return a newline-delimited list of the original feature names, followed by "->", followed by either NO CHANGE or a generalized feature name. For example

    feature\_a -> feature\_letter
    
    feature\_b -> feature\_letter
    
    feature\_letter -> feature\_letter
    
    feature\_1 -> NO CHANGE
\end{promptbox}

The following prompt applies once per feature $\phi_i$ in the current representation $\Phi^k$, as this yielded more consistent results than a prompt asking for the full mask at once.
\begin{promptbox}[Mask: Compute Mask Element $m_i$]
    This is an environment config: \textcolor[HTML]{F7941D}{\{environment\_description\}}\\

    The following is an instruction for a task that the agent needs to complete in this environment. instruction = \textcolor[HTML]{F7941D}{\{instruction\}}\\

    The following is a history of corrections that the agent has been provided. history = \textcolor[HTML]{F7941D}{\{corrections history\}}\\
    
    The following is an addendum to the original instruction after the agent asked for clarification about the task. addendum = \textcolor[HTML]{F7941D}{\{correction\}}\\

    The current available features are: \textcolor[HTML]{F7941D}{\{$\Phi^k$\}}\\

    We are evaluating whether \textcolor[HTML]{F7941D}{\{$\phi_i$\}}\\ is part of the minimal set of features needed for the agent to compute its reward given that it has NO access to the instruction, corrections, or addendum text -- only to the feature values. Please return either a 1 or 0 for the following feature: feature = \textcolor[HTML]{F7941D}{\{$\phi_i$\}}, where 1 means that the agent uses this feature to calculate its reward and 0 means that the agent does not need this feature when calculating its reward.\\

    Please reason step by step. In particular, consider whether the feature provides precise and relevant information for the task. Also consider whether the feature might lead the agent to treat multiple objects the same way, even if only one of them is relevant to the instruction.\\

    For example, for the instruction go to a book and the features ["book\_dist", "table\_dist", "chair\_dist"], the minimal set would include only the features that refer specifically to the object mentioned in the instruction.\\
    
    Provide your answer as ONLY a single integer. Do not include any surrounding explanation or text.
\end{promptbox}

\begin{promptbox}[Example Prompt Inputs]
    \textcolor[HTML]{F7941D}{environment description}: See "2D Navigation Example Environment Description" and "Franka 7DoF Example Environment Description" below \\

    \textcolor[HTML]{F7941D}{instruction}: "Move the full coffee cup" \\

    \textcolor[HTML]{F7941D}{corrections history}: ["Maybe try to swerve near the bowl?", "Don't swing the coffee so much.", "Don't go near the laptop."] \\

    \textcolor[HTML]{F7941D}{correction}: "Ignore what I said about the bowl, just avoid the laptop"\\

    \textcolor[HTML]{F7941D}{$\Phi^k$}: [distance\_to\_bowl, coffee\_orientation, laptop\_avoidance] \\
\end{promptbox}

\vspace{20em}

\begin{promptbox}[Add and Generalize: Generate Code]
        Based on the feature identified above, write a Python method that takes as input: \textcolor[HTML]{F7941D}{\{function inputs\}}\\

        In your method, you may use get\_objects\_with\_property(property: str, environment: dict, robot\_env), which takes a property name, the environment dict, and the robot environment object, and returns a list of [x, y, z] positions for all objects in the environment that have that property, type, or name. You must import this function from flexible\_features.utils.env\_utils.\\

        \textcolor[HTML]{F7941D}{\{example distance feature\}}\\
        
        \textcolor[HTML]{F7941D}{\{example orientation feature\}}\\

        Only include one feature (e.g. distance\_to\_book) and not a general feature (e.g. distance\_to\_object) in your answer. Do not use objects or properties that do not exist in this environment. Please write code that is generalizable to new environments (for example, the new environment may have objects in different positions than the current environment)\\
        
        Output your answer in form:
        
        Feature name: "feature\_name"
        
        Feature code:
        
        ```python
        
        import ...
        
        def feature\_name(environment, trajs, robot\_env):
        
            feat\_val = np.zeros((trajs.shape[0], 1))
            
            for pos in ...:
            
                ...
                
                feat\_val += ...  \# (K, 1)
                
            return feat\_val
            
        ```\\
        
        Please write code for all the features you have identified, and ensure the code is executable. Make sure your code is generalizable to new environments. Don't assume you are given any other information about the environment other than what is explicitly specified in the environment config. Do not include any surrounding explanation or text.
\end{promptbox}

\begin{promptbox}[2D Navigation System Prompt]
    An agent is in a 2D gridworld environment of size 9x9. (0,0) is the top left corner and (8, 8) is the bottom right corner (note this is not a typical coordinate grid).

    The environment config has the following attributes:
    
        \tab X (int): width of the environment
        
        \tab Y (int): height of the environment
        
        \tab starts (List[Tuple[int]]): starting coordinate of the agent
        
        \tab goals\_pos (List[Tuple[int]]): coordinate position of the goal
        
        \tab goals\_color (str): color of the goal
        
        \tab goals\_type (str): object type of the goal
        
        \tab objects\{1-3\}\_pos (List[Tuple[int]]): position(s) of other objects in the environment. This will be empty if there is no object in that slot.
        
        \tab objects\{1-3\}\_color (str): color of the corresponding object at the position objects\{1-3\}. This will be empty if there is no object in that slot.
        
        \tab objects\{1-3\}\_type (str): object type of the corresponding object at the position objects\{1-3\}. This will be empty if there is no object in that slot.
\end{promptbox}

\begin{promptbox}[Franka 7DoF System Prompt]
    A 7DoF Franka robot arm is carrying \textcolor[HTML]{F7941D}{a full coffee cup / a lit candle / a heavy bowling ball} on a tabletop environment. The arm is controlled via a PyBullet simulator.

    The environment config has the following attributes:

    \tab objects:

        \tab \tab object1:

            \tab \tab \tab urdf: object1.urdf
            
            \tab \tab \tab position: [x, y, z]
            
            \tab \tab \tab properties: [\textcolor[HTML]{F7941D}{object properties}]

        \tab \tab ...

    \tab \tab tabletop: z \# z coordinate plane of the tabletop
\end{promptbox}

\begin{promptbox}[2D Navigation Example Environment Description]
env:

  type: gridrobot
  
  X: 9
  
  Y: 9
  
  starts: [[0,0]]
  
  goals\_pos: [[8,8]]
  
  goals\_color: `green'
  
  goals\_type: `cross'
  
  objects1\_pos: [[0,8]]
  
  objects1\_color: `blue'
  
  objects1\_type: `fridge'
  
  objects1\_property:[`fridge', `light\_source', `icy', `stationary', `appliance']
  
  objects2\_pos: [[8,0]]
  
  objects2\_color: `red'
  
  objects2\_type: `stove'
  
  objects2\_property:[`stove', `heat\_source', `metal', `stationary', `appliance']
  
  objects3\_pos: [[4,4]]
  
  objects3\_color: `orange'
  
  objects3\_type: `kettle'
  
  objects3\_property:[`kettle', `portable', `heat\_source', `appliance']
  
\end{promptbox}

\begin{promptbox}[Franka 7DoF Example Environment Description]
env:

  objects:
  
    \tab laptop:
    
      \tab \tab urdf: laptop.urdf
      
      \tab \tab position: [-0.6, -0.5, 0.64]
            
      \tab \tab properties: ["laptop", "rectangular", "metal", "electronic"]
      
    \tab bowl:
    
      \tab \tab urdf: bowl.urdf
      
      \tab \tab position: [-0.6, -0.8, 0.9]
      
      \tab \tab properties: ["bowl", "wooden", "lightweight", "round", "dishware"]
      
    \tab phone:
    
      \tab \tab urdf: phone.urdf
      
      \tab \tab position: [-0.6, -0.5, 0.64]
            
      \tab \tab properties: ["phone", "lightweight", "metal", "electronic"]
      
\end{promptbox}

\begin{promptbox}[2D Navigation: Feature Function Inputs]
    Based on the feature description above, write a Python method that takes as input:
    
    * the ground truth environment features specified in the config (dict)
    
    * a trajectory (List[float]) representing the agent's position over time as a list of x,y coordinates (e.g. [0.0 0.0 0.0 1.0] means the agent started at (0,0) and moved to (0,1). The trajectory should be of length 34 (representing 17 timesteps where each timestep has a x,y coordinate for the agent's position)\\

    and outputs:
    
    * the cumulative value for the feature over the course of the entire trajectory (averaged over multiple objects if present) (array with dtype=np.float64).
    
\end{promptbox}

\begin{promptbox}[7DoF Franka: Feature Function Inputs]
    Based on the feature identified above, write a Python method that takes as input:
    
    * the ground truth environment features specified in the config (dict)
    
    * a batch of trajectories (dim KxTxN)
    
    * the robot environment object (robot\_env)\\
    
    and outputs:
    
    * the cumulative value for the feature over the course of all of the trajectories, shape (K, 1)\\
    
    Each trajectory batch is a KxTxN tensor where:
    
        K is the number of trajectories in the batch
        
        T is the number of timesteps (21)
        
        N is the state dimension\\
        
    Each state in the timestep is an N-dimensional vector laid out as follows:
    
    waypt[:7]               -- joint angles
    
    orientations.flatten()  -- robot link orientations
    
    coords.flatten()        -- robot link xyz coordinates
        
    The end-effector xy position is at indices 88:90 of each state vector.

\end{promptbox}

\begin{promptbox}[2D Navigation: Example Distance Feature]
    For example, for "agent distance to book objects" feature, you may write code\\
    
    '''from flexible\_features.utils.env\_utils import get\_objects\_with\_property\\

        def agent\_distance\_to\_book\_objects(environment, trajectory, r=3.0): 
        
        \# environment is the ground truth environment \\

        \tab \# returns the position(s) of the object(s) with property "book"
        
        \tab book\_objects\_pos = get\_objects\_with\_property("book", environment) \\

        \tab \# initialize the feature value
        
        \tab book\_objects\_feat = np.zeros(1) \\
        
        \tab \# iterate through the trajectory
        
        \tab for i in range(0, len(trajectory) - 1, 2): \\

        \tab \tab \# iterate through the book objects
            
        \tab \tab for book\_object in book\_objects\_pos: \\

        \tab \tab \tab \# calculate the dist between the agent and the book object (x)
        
        \tab \tab \tab dx = abs(trajectory[i] - book\_object[0]) \\

        \tab \tab \tab \# calc the distance between the agent and the book object (y)

        \tab \tab \tab dy = abs(trajectory[i + 1] - book\_object[1])
                
        \tab \tab \tab dist = np.linalg.norm([abs(dx), abs(dy)])\\ 
                
        \tab \tab \tab \# Cap distance at r (saturate)
                
        \tab \tab \tab bump\_dist = min(dist, r)
              
        \tab \tab \tab book\_objects\_feat += bump\_dist\\

        \tab \# return the average distance to the book object(s) for each trajectory

        \tab feature\_value = book\_objects\_feat / len(book\_objects\_pos) 
        
        \tab return feature\_value \# array with dtype=np.float64'''
\end{promptbox}

\begin{promptbox}[7DoF Franka: Example Distance Feature]
    For example, for "distance to book" feature, you may write code:\\
    
    '''import numpy as np
    
    from flexible\_features.utils.env\_utils import get\_objects\_with\_property\\

    def distance\_to\_book(environment, trajs, robot\_env, margin=0.4):\\
    \tab """
    
        \tab Computes cumulative dist reward from end-effector to book in xy coords.
        
        \tab Params:
        
            \tab \tab environment -- dict with 'objects' mapping object names to their data
            
            \tab \tab trajs       -- np.ndarray of shape (K, T, N), batch of K trajectories
            
            \tab \tab robot\_env   -- robot environment object for live object position lookup
                            
    \tab """\\

        \tab book\_positions = get\_objects\_with\_property("book", environment, robot\_env) 
        
        \tab feat\_val = np.zeros((trajs.shape[0], 1))
        
        \tab for pos in book\_positions:\\
        
            \tab \tab book\_xy = np.asarray(pos[:2])                           
            
            \tab \tab ee\_xy = trajs[:, :, 88:90]                              
            
            \tab \tab dists = np.linalg.norm(ee\_xy - book\_xy, axis=2) - margin
            
            \tab \tab feat\_val += np.sum(-((dists < 0) * dists), axis=1, keepdims=True)  \\
            
        \tab return feat\_val / max(1, len(book\_positions)) ''' \\
    
    feat\_val is a (K, 1) array representing the average over matched objects of the summed per-timestep distance reward across the trajectory batch.
\end{promptbox}

\begin{promptbox}[7DoF Franka: Example Orientation Feature]
    Another example, for "orientation of coffee" feature, you may write code:\\
    
    ```import numpy as np\\

    def coffee\_orientation\_feature(environment, trajs, robot\_env):\\
    
        \tab """Computes the coffee orientation feature by checking if the end effector is oriented vertically."""\\
        
        \tab \# Rotation matrix of link6 is stored as a flattened 3x3 at indices 61:70
        
        \tab link6\_rot\_flat = trajs[:, :, 61:70]                             
        
        \tab K, T, \_ = link6\_rot\_flat.shape
        
        \tab link6\_rot = link6\_rot\_flat.reshape(K, T, 3, 3)

        \tab \# The z-component of the EE x-axis indicates how much it points downward
        
        \tab \# Pulled directly from rotation matrix (no additional transform needed)
        
        \tab ee\_x\_axis\_z\_component = link6\_rot[:, :, 2, 0]                   
        
        \tab return np.sum(1 - ee\_x\_axis\_z\_component, axis=1, keepdims=True) 
        
    '''

\end{promptbox}

\subsection{Baseline Implementations}

\begin{promptbox}[MFE: Get Exhaustive Feature List]
    This is an environment config: \textcolor[HTML]{F7941D}{\{environment\_description\}}\\

    Use this to give specific feature(s) in that environment that explains what are relevant properties in the environment. If there are multiple of the same object, please give the average distance to all objects of that type.\\
    
    Please specify feature(s) that we can write code to compute from the position of the agent to other objects in the scene. This is the current list of features that we already know: \textcolor[HTML]{F7941D}{\{$\Phi^k$\}}. Please give different features in your answer (e.g. if distance to goal feature is already in the list, do not give distance to goal again).\\
    
    Please give your confidence for each feature between 0 and 1. You may ground your answer in visible objects in the scene, and give answers that we can compute relative to the position of the agent.
\end{promptbox}

\begin{promptbox}[USF: Get Updated Feature List]
    This is an environment config: \textcolor[HTML]{F7941D}{\{environment\_description\}}\\

    The following is an instruction for a task that the agent needs to complete in this environment. instruction = \textcolor[HTML]{F7941D}{\{instruction\}}\\

    The following is a history of corrections that the agent has been provided. history = \textcolor[HTML]{F7941D}{\{corrections history\}}\\
    
    The following is the most recent addendum to the original instruction after the agent asked for clarification about the task. addendum = \textcolor[HTML]{F7941D}{\{correction\}}\\

    Based on this, create a feature list that contains computable numeric values (e.g. "distance from agent to object X") derived from scene state. These features will be used by a downstream model that has NO access to the instruction, corrections, or addendum text -- only to the feature values.\\

    If there are multiple of the same object, please give the average distance to all objects of that type. Please specify feature(s) that we can write code to compute from the position of the agent to other objects in the scene. Every feature in your answer must be unique (e.g. if one feature is distance to goal, do not create another feature that is distance to goal again).Please give your confidence for each feature between 0 and 1. Ground your answer in visible objects in the scene, and give answers that we can compute relative to the position of the agent.
\end{promptbox}

MFE uses an equivalent masking prompt as above. MFE and USF use an equivalent code generation prompt as above. Our implementation and prompts for \textbf{ALGAE} exactly follow \cite{peng2024adaptive}.

\subsection{RQ1 Additional Details}
We evaluate the contribution of each operation by constructing seven scenarios that represent all combinations of \textsc{Add}, \textsc{Generalize}, and \textsc{Mask}. Table~\ref{tab:scenarios} lists the initial features and expected outputs for each scenario in the KitchenWorld domain with the correction ``Don't burn yourself on the stove.'' For example, in the \textsc{Mask}+\textsc{Generalize} scenario, the initial features measure distance to the goal, the stove, and the fridge. The correction indicates that the robot should not care about the fridge at all, so the fridge feature should get \textsc{Mask}ed away. Since a stove-related feature already exists, the \textsc{Add} operation is not necessary. However, the existing feature should be \textsc{Generalize}d to apply to all dangerous heat sources.

\begin{table}[h]
  \setlength{\belowcaptionskip}{-6pt}
  \centering
  \setlength{\tabcolsep}{4pt}
  \begin{tabular}{lll}
    \toprule
    \textbf{Operations} & \textbf{Initial Rep} & \textbf{Expected Actions} \\
    \midrule
    Add                & goal                      & add stove \\
    Generalize         & goal, stove               & generalize stove $\to$ heat source \\
    Mask               & goal, stove, fridge       & mask fridge \\
    Add + Generalize   & goal, fridge              & add stove, generalize stove $\to$ heat source \\
    Add + Mask         & goal, fridge              & add stove, mask fridge \\
    Mask + Generalize  & goal, stove, fridge       & mask fridge, generalize stove $\to$ heat source \\
    All Three          & goal, fridge              & add stove, mask fridge, generalize stove $\to$ heat source \\
    \bottomrule
    \addlinespace[4pt]
  \end{tabular}
  \caption{2D Navigation misalignment scenarios. Each scenario requires a distinct subset of \ours's three operations. The initial features determine which operations are needed to recover the true reward, and the expected actions describe the correct response to the language correction ``Don't burn yourself on the stove.'' All features are Euclidean distances to the named object or location.\label{tab:scenarios}}
  \vspace{-1em}
\end{table}

To quantify per-operation reliability, we ran \textsc{Add}, \textsc{Generalize}, and \textsc{Mask} in isolation, outside the full \ours pipeline, on all seven 2D Navigation scenarios. We used 15 additional seeds, separate from the five evaluation seeds, yielding 105 trials. Each operation received the scenario's initial misaligned feature set and natural language correction as input. A trial counts as an error if the operation's output differs from the expected edit, either because it (i) applied an edit when none was warranted (or omitted one that was), or (ii) applied the edit but produced the wrong feature(s). Error rates were 8.6\% for \textsc{Add} (9/105), 3.8\% for \textsc{Mask} (4/105), and 5.7\% for \textsc{Generalize} (6/105).

\subsection{RQ2 Additional Details}
To validate \ours performance in more realistic settings, we design a sequence of five object configurations in the Franka 7DoF tabletop manipulation domain through which each operation of \ours becomes necessary. Fig.~\ref{fig:franka_seq} lists an example sequence of corrections in the CoffeeRobot task variant. Each stage of the sequence depicts the object configuration, an incorrect robot trajectory (lavender), the user's ground-truth trajectory (green), an example language correction, a potential representation update, and the resulting aligned trajectory for that specific configuration (purple). Throughout the example CoffeeRobot sequence, the user's ground-truth preference is to avoid any damage from spilling coffee on their laptop, phone, calculator, or other electronic device.

\noindent{\textbf{Stage 1:}} The user only sees a bowl and, mistakenly believing it to be relevant, provides the correction ``Go near the bowl." In response, the robot \textsc{Add}s a new feature related to bowl proximity, establishing a use case where an incorrect or irrelevant feature enters the robot's representation.

\noindent{\textbf{Stage 2:}} The user sees a laptop in addition to the bowl and wants the robot to keep the coffee far away from the laptop with the correction ``Don't spill on my laptop." To accommodate the user's stated preference, the robot must \textsc{Add} a new feature related to laptop avoidance.

\noindent{\textbf{Stage 3:}} The user sees the bowl overlap with the laptop. However, the robot's previously learned bowl proximity feature becomes a spurious correlation, distorting the trajectory by simultaneously pulling it away from the laptop and towards the bowl. Since the user truly doesn't care about the bowl, they give the correction ``Wait, ignore the bowl now," causing the robot to \textsc{Mask} away this now-irrelevant feature.

\noindent{\textbf{Stage 4:}} The user sees a phone overlap with the laptop, and desires to keep the coffee far away from both devices. Even with only a laptop avoidance feature, the robot maintains aligned behavior, because a trajectory that veers away from the laptop also veers away from the overlapping phone. Visually, the user believes that little to no correction is necessary: ``Good, don't spill on my laptop.'' Although the robot's representation does not change, this stage demonstrates that behavioral alignment can mask representational misalignment. The robot \textit{appears} to be avoiding both the laptop and the phone, but its internal representation reveals that the robot does not consider the phone at all. This hidden misalignment will only become apparent when the laptop is removed.

\noindent{\textbf{Stage 5:}} Now the laptop is removed, exposing the misalignment from Stage 4. The robot's laptop avoidance feature is ineffective with only a phone present. The user provides a correction ``Don't spill on my phone too,'' and the robot must \textsc{Generalize} its learned laptop avoidance feature to all electronics devices. Only then can the robot maintain aligned behavior in environments with only a laptop, a phone, or any other electronic device.

Together, these five stages show that behavioral alignment is necessary but not sufficient to guarantee robustness to new configurations. Despite a correction producing aligned behavior on every stage, a new configuration quickly reveals further misalignments that must be combated through a combination of \textsc{Add}, \textsc{Generalize}, and \textsc{Mask} operations.

\subsection{RQ3 Additional Details}
\label{app:likert}

\noindent\textbf{User Study Interface.} Participants viewed the robot trajectories through a simulated interface consisting of a main screen showing the scene and robot, and a side panel containing the available controls. They could replay the goal and current trajectories as many times as they wished, and typed their natural language corrections for each stage. To inspect the scene from different angles, they could switch between three preset camera views (``Top Down,'' ``Side,'' and ``Front''), and the tutorial additionally taught them to rotate, pan, and zoom the scene with the computer mouse. Although the drawn-out trajectories were not visible to participants, we visualize them here to illustrate how the goal and current trajectories can differ.

\begin{figure}[h!]
    \centering
    \includegraphics[width=0.75\linewidth]{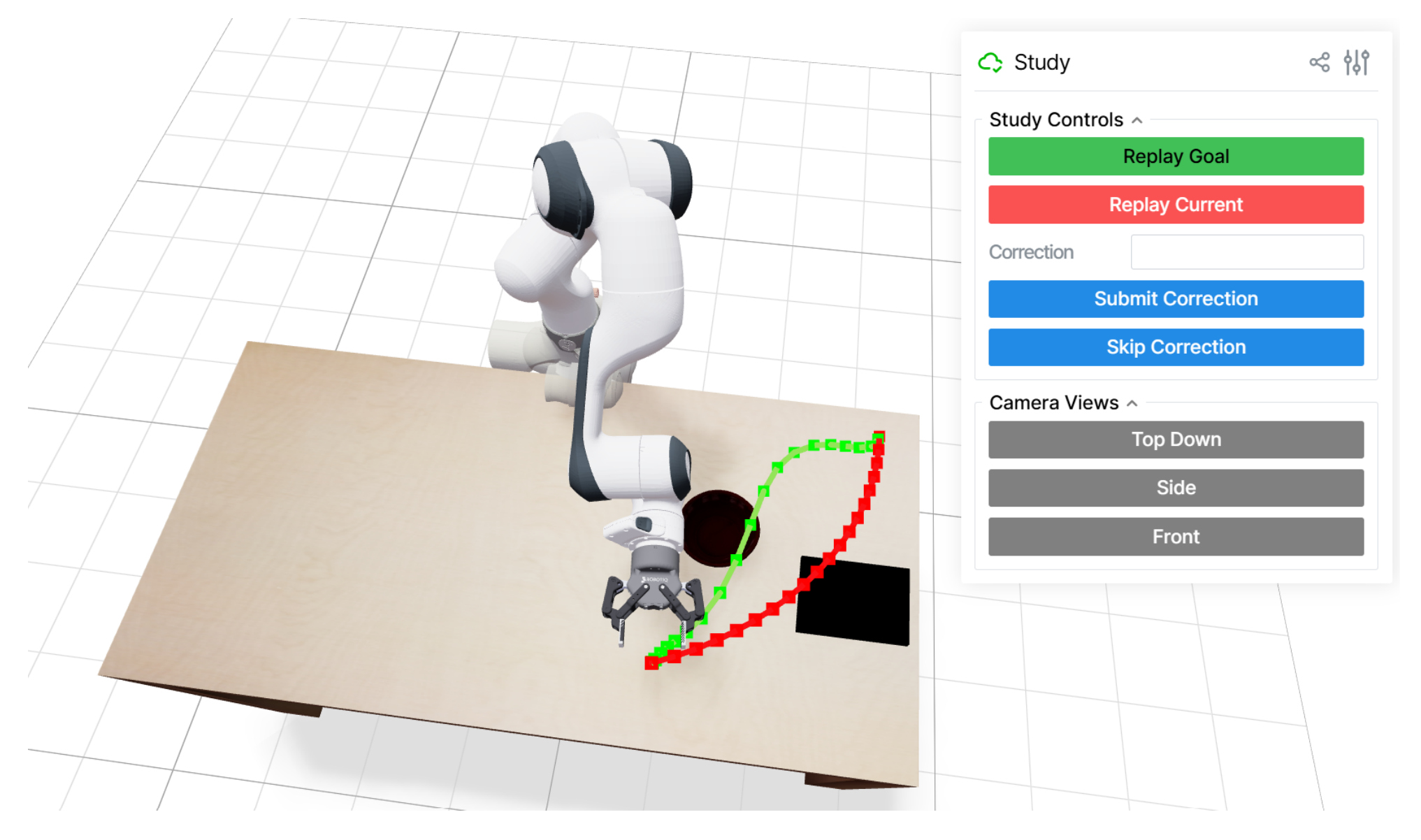}
    \captionof{figure}{%
    User study interface. The side panel includes buttons to replay the ground truth (green) and current (red) trajectories, as well as pre-set camera views labeled ``Top Down,'' ``Side,'' and ``Front.'' Participants enter their natural language corrections in the text field on the side panel. Participants were also instructed on how to use the computer mouse to rotate, pan, and zoom the scene.
     \label{fig:user_interface}}
\end{figure}

\noindent\textbf{Likert Items.} Please rate your agreement with each statement about the robot learning via Algorithm 1/2/3. 

\textit{Strongly disagree (1) to Strongly agree (7)}
\begin{enumerate}
    \item Collaborativity Scale
    \begin{itemize}
        \item The robot did not collaborate with me to complete the task.
        \item The robot worked with me to complete the task.
    \end{itemize}
    \item Ease of Use Scale
    \begin{itemize}
        \item The robot required minimal corrections.
        \item I had to keep correcting the robot.
    \end{itemize}
    \item Task Performance Scale
    \begin{itemize}
        \item The robot accomplished the task in the way I wanted.
        \item The robot was NOT able to complete the task correctly.
    \end{itemize}
    \item Understanding of User Intent Scale
    \begin{itemize}
        \item The robot understood how I wanted the task to be done.
        \item The robot learned from my corrections.
    \end{itemize}
    \item Robustness to Unintended Consequences Scale
    \begin{itemize}
        \item The robot wrongly updated its understanding about aspects of the task I did NOT want to change.
        \item After I adjusted the robot, it continued to perform the other parts of the task correctly.
        \item After I adjusted the robot, it incorrectly updated parts of the task that were already correct.
    \end{itemize}
\end{enumerate}

\noindent Please write any notes or observations you have about Algorithm 1/2/3.

\noindent What preferences were you trying to get the robot to learn in Task 1/2/3?

\noindent Please write any observations you have about the differences between Algorithms 1, 2, and 3.

\end{document}